\documentclass[conference]{ieeeconf}
\IEEEoverridecommandlockouts 
\usepackage{graphicx}
\usepackage{algorithm}
\usepackage{algpseudocode}
\usepackage{amsmath} 
\usepackage{xcolor}
\usepackage{comment}

\usepackage{microtype}
\usepackage{booktabs}
\usepackage{multirow}
\usepackage{hyperref}
\usepackage[caption=false]{subfig}

\usepackage{amssymb}
\usepackage{mathtools}
\usepackage{graphicx}

\hypersetup{
    colorlinks,
    linkcolor={blue!50!black},
    citecolor={blue!50!black},
    urlcolor={blue!80!black}
}
\usepackage[textsize=tiny]{todonotes}

\title{\LARGE \bf
Option Discovery Using LLM-guided\\ Semantic Hierarchical Reinforcement Learning
}
\author{Chak Lam Shek$^{1}$ and Pratap Tokekar$^{2}$% <-this % stops a space
\thanks{This work has been submitted to the IEEE for possible publication. 
Copyright may be transferred without notice, after which this version may no longer be accessible. 
This is the preprint version.}
\thanks{$^{1}$Department of Electrical Engineering, University of Maryland, College Park, USA
        {\tt\small cshek1@umd.edu}}%
\thanks{$^{2}$Department of Computer Science, University of Maryland, College Park, USA
        {\tt\small tokekar@umd.edu }}%
}

\begin{document}

\maketitle
\thispagestyle{empty}
\pagestyle{empty}

\begin{abstract}
Large Language Models (LLMs) have shown remarkable promise in reasoning and decision-making, yet their integration with Reinforcement Learning (RL) for complex robotic tasks remains underexplored. In this paper, we propose an LLM-guided hierarchical RL framework, termed LDSC, that leverages LLM-driven subgoal selection and option reuse to enhance sample efficiency, generalization, and multi-task adaptability. Traditional RL methods often suffer from inefficient exploration and high computational cost. Hierarchical RL helps with these challenges, but existing methods often fail to reuse options effectively when faced with new tasks. To address these limitations, we introduce a three-stage framework that uses LLMs for subgoal generation given natural language description of the task, a reusable option learning and selection method, and an action-level policy, enabling more effective decision-making across diverse tasks. By incorporating LLMs for subgoal prediction and policy guidance, our approach improves exploration efficiency and enhances learning performance. On average, LDSC outperforms the baseline by 55.9\% in average reward, demonstrating its effectiveness in complex RL settings. More details and experiment videos could be found in \href{https://raaslab.org/projects/LDSC/}{this link\footnote{https://raaslab.org/projects/LDSC}}.
\end{abstract}

\section{Introduction}
In many decision-making scenarios, robots are tasked with efficiently managing a wide array of tasks, each exhibiting varying levels of complexity \cite{liu2023mtd, fu2024multi}. Despite significant advancements in Reinforcement Learning (RL), several challenges remain, including inefficient exploration  \cite{mnih2013playing, burda2018exploration} and high computational cost \cite{schulman2015trust, henderson2018deep}, particularly when robots are required to adapt to new goals or environments. These challenges stem from the need to repeatedly explore similar state spaces across different tasks, which makes the learning process both time-consuming and resource-intensive. To address this issue, some approaches (e.g., \cite{tao2021repaint}), advocate for transferring pretrained policies across various environments. However, such pretrained policies often fail to generalize effectively, offering minimal benefits in completely new tasks or when prior experience is insufficient \cite{dhiman2018critical, rusu2015policy}.

\begin{figure}
    \centering
    \includegraphics[width=0.7\linewidth]{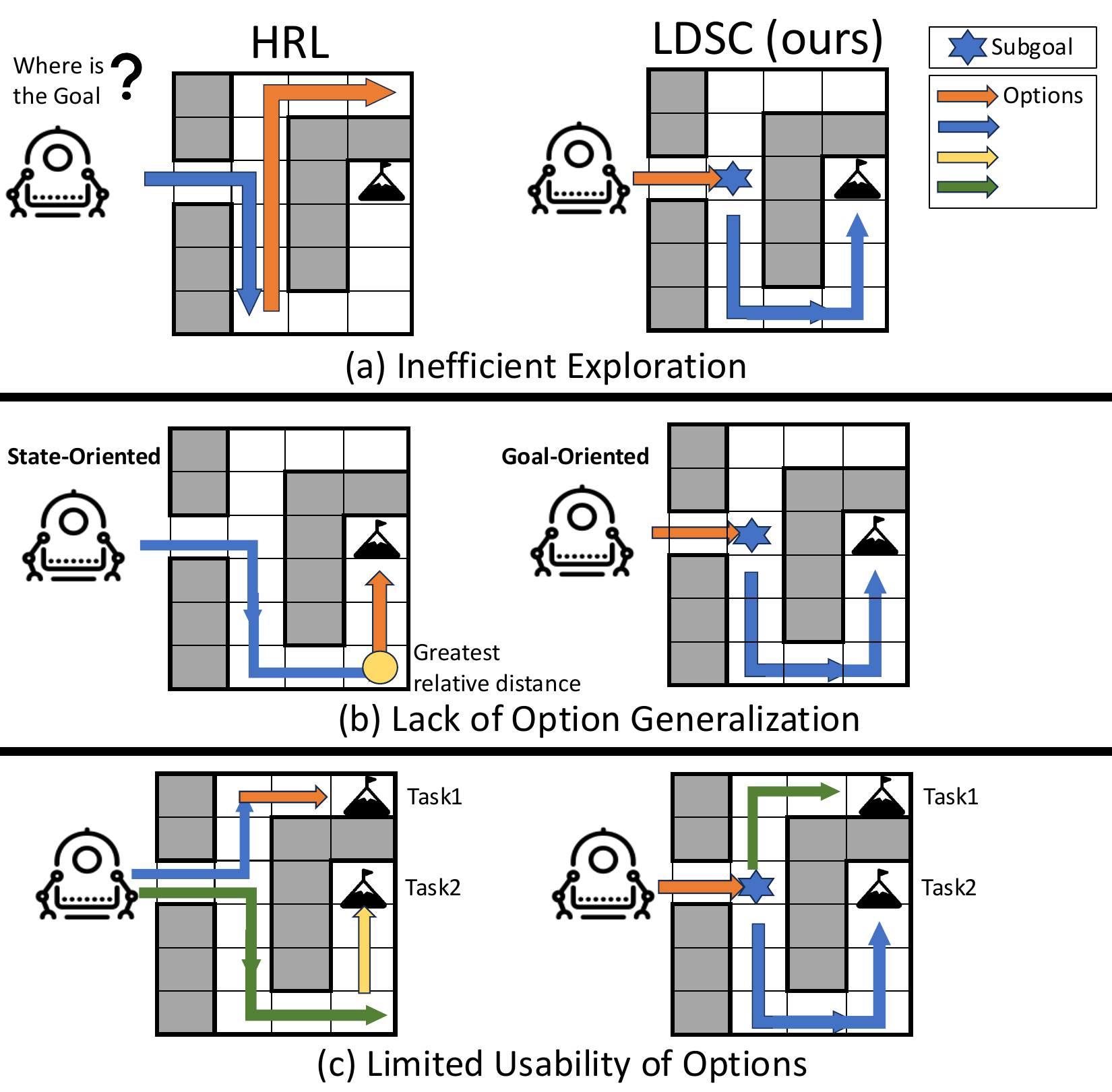}
    \caption{Comparison of HRL and LDSC in terms of exploration, generalization, and usability. (a) Inefficient Exploration: LDSC improves exploration by leveraging subgoals to guide the agent efficiently, reducing the need for exhaustive search. (b) Lack of Option Generalization: Unlike HRL, which learns state-oriented options that do not generalize well, LDSC learns goal-oriented options that can be reused across different tasks. (c) Limited Usability of Options: LDSC structures options around subgoals, enhancing their reusability for multiple tasks, improving learning efficiency, and enabling better skill transfer.}
    \label{fig:Comparsion}
\end{figure}

Numerous studies have investigated Hierarchical Reinforcement Learning (HRL) \cite{barto2003recent}, which aims to learn reusable options \cite{sutton1999between} or skills \cite{thrun1994finding} from tasks and apply them to new problem settings. One prominent example is the Deep Skill Chaining (DSC) \cite{bagaria2019option}, which hierarchically decomposes decision-making into a policy over options \footnote{An \textit{option} is defined as a tuple $(\mathcal{I}, \pi, \beta)$, consisting of an initiation set $\mathcal{I}$, an intra-option policy $\pi$, and a termination condition $\beta$ \cite{NIPS1994_07871915}.} for skill selection and intra-option policies for generating low-level actions. The success of this approach lies in grouping actions into options that can be efficiently selected and executed to achieve long-term goals. However, HRL encounters three primary challenges, as illustrated in Figure \ref{fig:Comparsion}: 1) \textbf{Inefficient Exploration} where robots face difficulties in developing effective option policies for long-horizon tasks \cite{konidaris2009skill} due to the need to reach goals before receiving positive feedback, leading to slow learning progress. 2) \textbf{Limited Policy Generalization} as most HRL approaches are heavily state-dependent \cite{machado2017eigenoption}, causing learned policies to be difficult to abstract or transfer to tasks with different state configurations. 3) \textbf{Restricted Usability of Learned Options} as these options are often tied to specific state transitions or distances, limiting their ability to be reused across tasks with varying goals or environments. Collectively, these challenges significantly impede HRL’s capacity to generalize and adapt to new and diverse settings.

Recent advancements in decision-making have explored reasoning-based approaches, such as leveraging Large Language Models (LLMs) to generate plans \cite{liu2023llm} or provide rewards \cite{kwon2023reward} that guide robot behavior. These approaches integrate Semantic reasoning to structure decision-making, enabling robots to plan and adapt to new tasks more efficiently. By employing semantic representations, robots can reason about abstract concepts such as subgoals and actions, which is particularly beneficial when facing tasks with high complexity or unfamiliar environments.

To address the challenge of reusing option policies in multi-task settings, we propose LLM-guided Semantic Deep Skill Chaining (LDSC), a Semantic HRL method that leverages semantic logic to enable effective policy transfer across different tasks. The method follows a three-level hierarchy: the subgoal policy operates at a reasoning level, determining the sequence of subgoals required to achieve the high-level goal. This creates a logical framework that guides the option and action policies. The option policy selects the appropriate option for the chosen subgoal, while the action policy generates the corresponding low-level actions.

LDSC leverages LLM reasoning to guide the learning process and addresses several challenges in traditional HRL. By enabling the robot to reason about subgoals, LDSC provides a high-level, structured plan that alleviates the issue of inefficient exploration. Rather than relying on relative state distances, LDSC uses semantic representations that make the option policies more flexible and goal-oriented. This approach allows the robot to move meaningfully between subgoals, improving the reusability of learned options. Consequently, LDSC reduces the need for retraining and ensures that the learned options are adaptable and transferable across a variety of tasks, making it well-suited for complex, multi-task environments.

Below, we outline the three main contributions of this work:
\begin{itemize}
    \item \textbf{Improved Learning Efficiency:} Through LLM-generated subgoals, robots can achieve structured task completion, \textbf{accelerating the learning process}.
    \item \textbf{Policy Generalization:} Our approach enables effective transfer of learned policies across \textbf{diverse tasks}, promoting \textbf{generalization} and \textbf{adaptability} in complex environments.
    \item \textbf{Experimental Validation:} We demonstrate the effectiveness of our method through extensive experiments in diverse environments, showcasing its applicability to real-world multi-task challenges. On average, \textbf{LDSC outperforms baseline methods by 55.9\%}, reduces task completion time by \textbf{53.1\%}, and improves success rates by \textbf{72.7\%}, while maintaining a similar training time to DSC.

\end{itemize}

\section{Related Work} Sharma et al.~\cite{sharma2019dynamics} uses RL to discover low-level skills with predictable outcomes, allowing model-based planning in skill space. Bacon et al.~\cite{bacon2017option} introduces an option-critic architecture that autonomously learns temporal abstractions (options) in RL. Chunduru et al.~\cite{chunduru2022attention} proposes an attention-based extension that improves option diversity. Option discovery based on the construction of reusable skills through Successor Representations \cite{inproceedings}, learning from small fragments of experience within options \cite{ramesh2019successoroptionsoptiondiscovery}, the discovery of high-level behaviors from low-level actions \cite{fox2017multi}, unsupervised skill emergence from information-theoretic objectives  \cite{eysenbach2018diversity}, and the use of representations encoding diffusive information flow \cite{machado2017eigenoption}. Skill chaining \cite{konidaris2009skill} is a method for discovering new skills by creating chains of skills. Deep Skill Chaining \cite{bagaria2019option}, an extension of skill chaining enables the discovery of useful skills in high-dimensional spaces. In a related direction, \cite{ghugare2024closing} studies combinatorial generalization in RL and proposes temporal data augmentation to enable policy generalization across unseen state-goal pairs

Recently, numerous papers have explored the integration of LLMs for reasoning and their application in improving RL. Recent studies have demonstrated the use of LLMs for task decomposition into subproblems, including \cite{huang2022language, zhou2022least, pan2024hierarchical}. LLM-Planner \cite{song2023llm, liu2023llm} leverages LLMs to generate and adapt high-level plans using natural language instructions and environmental observations. Chain-of-thought methods have been explored to improve reasoning in LLMs \cite{wei2022chain, wang2023plan}. The use of tree structures for planning has also been studied in works such as \cite{zhou2023language, hu2023tree, yao2024tree}. Several works have explored using LLMs to convert natural language instructions into reward signals, including \cite{yu2023language, Li_2024_CVPR}. The CLIP model has been applied as a zero-shot reward model, assigning rewards based on a threshold \cite{rocamonde2023vision}. Language models have also been used to fine-tune offline RL tasks \cite{reid2022can} and to scaffold general sequential decision-making by initializing policies \cite{li2022pre}.

While prior works in HRL focus on skill discovery and option learning, and recent methods leverage LLMs for task decomposition and high-level planning, these approaches remain limited in their ability to automatically construct reusable, transferable skills for long-horizon tasks. Our work bridges these two directions by integrating LLM-guided subgoal discovery into the skill chaining framework, enabling the incremental construction of a hierarchy of generalizable options across diverse tasks.

\section{Problem Formulation}
In this work, we address a general decision-making problem where the tasks for the agent are provided as natural language, denoted by $ \{l_1, l_2, \dots, l_n\}$. Each $l_i$ represents a task instruction. These task instructions are grounded through \textit{goals}, defined as specific states the robot must reach to complete the task. The tasks are modeled as a semi-Markov Decision Process (semi-MDP) \cite{NIPS1994_07871915}, defined by the tuple $ \mathcal{M} = (\mathcal{S}, \mathcal{A}, P, R, \tau, \gamma) $, where $ \mathcal{S} $ is the set of states, $ \mathcal{A} $ is the set of actions, $ P(s'|s, a) $ is the probability of transitioning from state $ s $ to state $ s' $ after taking action $ a $, $ R(s, a) $ is the reward function, $ \tau(s, a) $ is the time duration associated with action $ a $, and $ \gamma $ is the discount factor. In this more general problem, the human language instructions $ \{l_1, l_2, \dots, l_n\} $ specify high-level tasks, which may be complex or abstract. The robot should interpret these instructions and decompose them into a sequence of smaller subgoals that can be more easily achieved. Each subgoal corresponds to a set of states  that satisfy the requirements of the subgoal within the semi-MDP, and the robot must develop a policy $ \pi $ that allows it to accomplish these tasks in a way that ultimately fulfills the overall objectives specified by $ \{l_1, l_2, \dots, l_n\} $.

For a particular task $l$, the robot seeks to optimize its policy $ \pi $ to maximize the expected cumulative reward by following the optimal sequence of actions that lead to the completion of the main goal. Formally, the objective is to find the policy $ \pi^* $ that maximizes the expected discounted reward: 

\begin{equation}
    \pi^* = \arg\max_{\pi} \mathbb{E} \left[ \sum_{t=0}^{\infty} \gamma^t R(s_t, a_t) \right]
\end{equation}
where $ \pi(a|s) $ is the policy that maps states to actions and $ R(s_t, a_t) $ is the reward at time $ t $. By breaking down complex goals into subgoals and solving them incrementally, the robot can handle more general tasks provided by natural language instructions.

\section{Our approach: LDSC}
\begin{figure}
    \centering
    \includegraphics[width=0.99\linewidth]{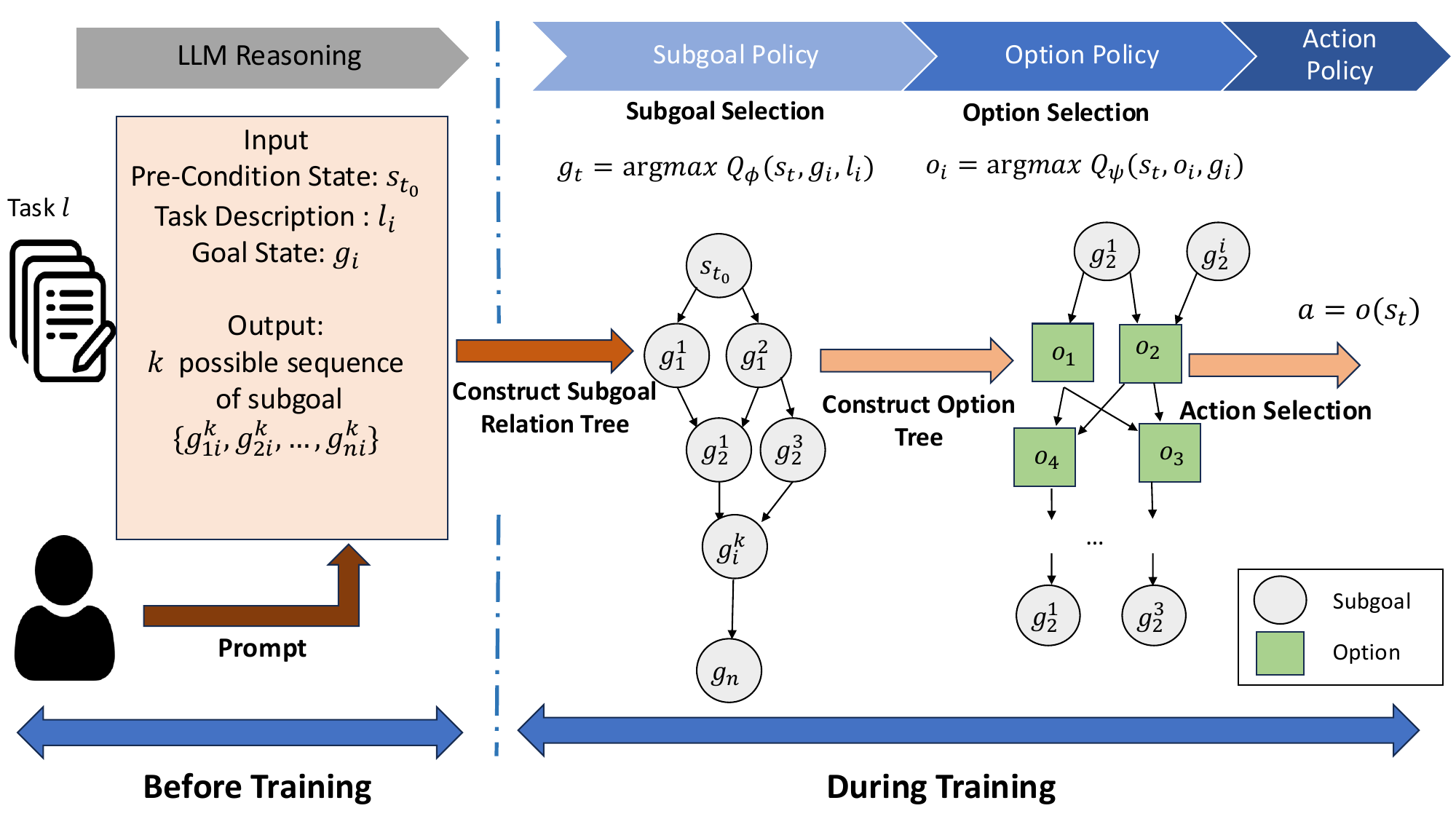}
    \caption{Overview of the \textbf{LDSC} framework. The process consists of two phases: Before Training and During Training. In the Before Training phase, an LLM generates subgoal sequences based on task descriptions and initial conditions. During Training, a hierarchical structure operates with three components: (1) the subgoal policy, which constructs a subgoal relation tree and selects subgoals; (2) the option policy, which builds an option tree and determines the best option; and (3) the action policy, which selects the final action based on the chosen option.}
    \label{fig:flow}
\end{figure}

Our approach, LDSC, leverages reasoning to enhance learning efficiency by breaking tasks into manageable subgoals. This hierarchical structure enables the robot to learn a smaller, reusable set of options that can be applied across multiple subgoals. LDSC operates on three levels: the subgoal policy, which oversees high-level task planning and subgoal selection; the option policy, which operates at the intermediate level by selecting and executing the appropriate option based on the chosen subgoal; and the action policy, which handles detailed actions required to complete each subgoal. 

\subsection{LLM Reasoning}
The algorithm begins with a human-provided instruction set $\{l_1, l_2, \dots, l_n\}$, which consists of a high-level goal, a description of the environment, constraints, and the initial state. Achieving such goals directly in sparse and complex environments is often difficult due to the large state space and the long horizons required to reach the goal. To address this challenge, the algorithm leverages the reasoning capabilities of a LLM to decompose the given instruction set into a series of manageable subgoals $\{g_{11}, g_{12}, \dots, g_{mn}\}$. Formally, the LLM decouples the original goal into a sequence of subgoals:

\begin{equation}
    G_n: \{g_{1i}, g_{2i}, \dots, g_{ji}\} = \text{LLM}(l_i, s_{t_0})
    \label{eq:LLM}
\end{equation}
where each $g_{ji}$ represents a potential intermediate step $j$ toward achieving the task $i$. This transformation allows the robot to focus on achieving smaller, more tractable goals in sequence, which are easier to accomplish than the original complex goal.

\begin{figure}
    \centering
    \includegraphics[trim=10 25 30 5, clip, width=0.8\linewidth]{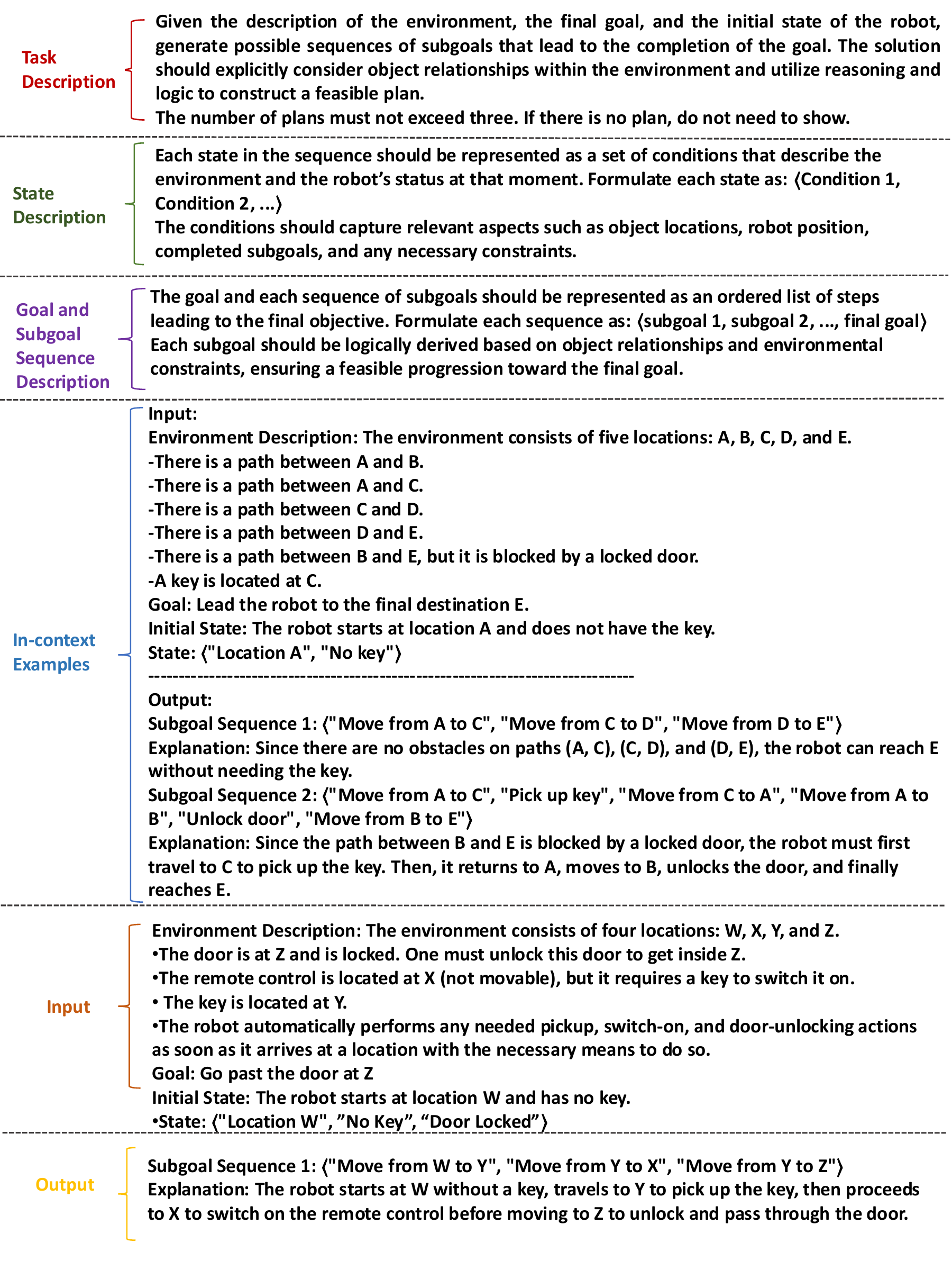}
    \caption{Structured prompt design for generating subgoal sequences in a robotic planning task. The figure outlines different sections of the prompt, including task description, state representation, goal and subgoal sequencing, and in-context examples. The example shown serves as a qualitative example based on the Point Maze environment, demonstrating how a robot navigates the maze by reasoning over object relationships and environmental constraints to generate feasible and logically ordered action sequences.}
    \label{fig:prompt}
\end{figure}

\subsubsection{Subgoal Relation Tree}

While LLMs possess strong reasoning capabilities, the subgoal sequences they suggest may not always be accurate or optimized. To address this, we draw inspiration from the Tree of Thought (ToT) framework \cite{yao2024tree} and propose constructing a subgoal relation tree to systematically explore and refine $k$ possible subgoal sequences $\{G_n^1, G_n^2 ..., G_n^k\}$ for task $n$. The tree is initialized with the current state $s_{t_0}$ as the root. From this root, the LLM generates multiple subgoal sequences, where each sequence represents a potential path toward achieving the final task.

The generated subgoal sequences are directly connected to the root state $ s_{t_0} $, forming the first layer of the subgoal relation tree. Specifically, each initial subgoal is defined as  

\begin{equation}
    g_{1i} = \text{create\_child\_subgoal}(s_{t_0})
    \label{eq:GoalTree}
\end{equation}
where $ g_{1i} $ represents the $1$st subgoal for task $ i $ in the first layer. Subgoals that appear in multiple sequences are treated as shared nodes within the tree, enabling efficient representation of overlapping paths.  

To expand the tree, for each subgoal $ g_{ji} $ in the current layer $ j $, we generate a new set of subgoals for the next layer $ j+1 $ using the recursive relation  
\begin{equation}
    g_{(j+1)i} = \text{create\_child\_subgoal}(g_{ji}), \quad \text{if } (g_{ji}, g_{(j+1)i}) \notin E
\end{equation}
where $ g_{(j+1)i} $ represents a newly generated child subgoal of $ g_{ji} $, and $ E $ denotes the set of existing edges in the tree. The expansion process continues recursively until one of the stopping criteria is met, such as a subgoal reaching the final task objective or the number of child nodes per subgoal exceeding a predefined threshold. 
This hierarchical structure enables systematic exploration of diverse task decompositions while ensuring that shared subgoals are efficiently reused across sequences. By leveraging this structured representation, planning efficiency is significantly improved for complex tasks.

\subsection{Subgoal Policy} 
After constructing the the relation tree, the algorithm proceeds to the training phase without further queries to the LLM. Next, the robot dynamically selects and evaluates goals based on  based on the set of next goals in the relation tree. At each state $s_t$, the robot selects the goal $g_t$ that maximizes its goal-value function. Since the robot must choose from a discrete set of goals, we employ Deep Q-learning (DQN) \cite{mnih2013playing} to learn the goal-value function:  
\begin{equation}  
g_t = \arg\max_{g_i \in G} Q_\phi(s_t, g_i, l).  
\label{eq:GoalControl}
\end{equation}  

The goal-value function is updated using the Semi-Markov Decision Process (SMDP) Q-learning update \cite{NIPS1994_07871915}. Given an SMDP transition $(s_t, g_t, r_{t:t+\tau}, s_{t+\tau})$, the robot learns to improve its policy by optimizing the Q-value associated with selecting a goal $g_t$ in state $s_t$. The target for updating the DQN, with Q-values parameterized by $\phi$, is defined as:

\begin{equation}  
y_t = \sum_{t'=t}^{t+\tau-1} \gamma^{t'-t} r_{t'} + \gamma^{\tau} Q_{\phi'}(s_{t+\tau}, \arg\max_{g' \in G} Q_\phi(s_{t+\tau}, g', l), l).  
\label{eq:UpdateG}
\end{equation}  

This formulation ensures that the robot can systematically select goals that maximize long-term rewards while accounting for the dynamics of the environment. By learning Q-values for state-goal pairs, the robot can prioritize goals that are most likely to achieve desired outcomes.

\subsection{Option Policy}
To lead the robot to the chosen subgoal, we choose $o_t$ according to $\pi_O(s_t)$ using Equations \ref{eq:option_condition} and \ref{eq:OptionControl}. The option $o_t$ is a temporally extended action that guides the robot through a sequence of actions to complete the subgoal.
We can learn its option-value function using Deep Q-learning (DQN):

\begin{equation}
O'(s_t) = \{o_i \mid I_{o_i}(s_t) = 1 \cap \beta_{o_i}(s_t) = 0, \forall o_i \in O \}
\label{eq:option_condition}
\end{equation}
\begin{equation}
o_t = \arg \max_{o_i \in O'(s_t)} Q_{\psi}(s_t, o_i, g)
\label{eq:OptionControl}
\end{equation}

Once the subgoals are defined, the robot employs a DSC approach to achieve each subgoal. At each step, the robot picks an option $o_t$ based on its current policy $\pi_{\mathcal{O}}(s_t, g_t)$. The environment is updated by executing the option, resulting in the reward $r_{t:t+\tau}$ and a new state $s_{t+\tau}$. After each option execution, the robot updates its policy using the collected data:
\begin{equation}
    \pi_{\mathcal{O}} = \text{update}(s_t, g_t, o_t, r_{t:t+\tau}, s_{t+\tau})
\end{equation}

The Q-value target for learning the weights $\psi$ of the DQN is given by:
\begin{equation}
\begin{aligned}
y_t = \sum_{t' = t}^{\tau} \gamma^{t' - t} r_{t'} +  \gamma^{\tau - t} Q_{\psi'}(s_{t + \tau}, \\
\arg \max_{o' \in O'(s_{t + \tau})} Q_{\psi}(s_{t + \tau}, o', g), g)
\end{aligned}
\label{eq:updateO}
\end{equation}
\begin{algorithm}
\caption{LDSC}
\small
\begin{algorithmic}

\State \textbf{Given} Human provides instruction set $ L = \{l_1, l_2, \dots, l_n\}$, hyperparameter $T_0$, the time budget of execution
\State \textbf{Robot’s Option Repertoire:} $\mathcal{O} = \{ \}$
\State \textbf{Policy over tasks:} $\pi_{\mathcal{O'}}: s_t \times l_i \to g_t$
\State \textbf{Untrained Options:} $o_U = \{\}$
\State \textbf{Policy over options:} $\pi_{\mathcal{O}}: s_t \times g_i \to o_t$
\State \textbf{Buffer:} $B = \{\}$
\State 
\For{ all $l_i$ in $L$}
    \State Use LLM to decouple instruction set into subgoals Set $\{G_i^1, G_i^2 ..., G_i^k\}$ using Equation \ref{eq:LLM}
    \State Construct the Subgoal Relation Tree using Equation \ref{eq:GoalTree}

    \For {all new subgal $g_i$}
        \State \textbf{Global Option:} $o_{z_i} = (I_{o_{z_i}}, \pi_{o_{z_i}}, \beta_{o_{z_i}} = 1_{g_i}, T = 1 )$
        \State \textbf{Goal Option:} $o_{g_i} = (I_{o_{g_i}}, \pi_{o_{g_i}}, \beta_{o_{g_i}} = 1_{g_i}, T = T_0)$
        \State Add $o_{g_i}$ into $o_U$
    \EndFor
\EndFor 
\For{ task $l_i$}
    \State  $s_t = s_0$
    
    \While{$t$ is not terminated}
        \State Choose a $g$ using equation \ref{eq:GoalControl}
        \State Choose a $o$ using equation \ref{eq:OptionControl}
        \State $r_{t:\tau}, s_{t+\tau} = execute\_option(o_t)$
        \State $\pi_{\mathcal{O}}$ = update($s_t$, $g$, $o_t$, $r_{t:t+\tau}$, $s_{t+\tau}$) using Equation \ref{eq:updateO}

        \State Buffer $B = B \cup r_{t:\tau}, s_{t+\tau}$
        \If{$s_{t+1}$ meet $g$}
            \State load $\hat{r}, \hat{s}$ in B
            \State  $\pi_{\mathcal{O'}}$ = update($s_t$, $g$,  $\hat{r}$, $\hat{s}$) using Equation \ref{eq:UpdateG}
            \State Buffer $B = \{\}$
        \EndIf
        \If{$\beta_{o_k}(s_{t+\tau})$ \textbf{and} $(s_0 \notin I_{o_i} \forall o_i \in \mathcal{O})$} , $o_k \in o_U$
            \State $o_k$.learn\_initiation\_classifier()
            \If{$o_k$.initiation\_classifier.is\_trained()}
                \State $\pi_{\mathcal{O}}$.add($o_k$) 
                \State $\mathcal{O'}$.append($o_k$)
                \State Construct the Option Tree using Equation \ref{eq:OptionTree}
            \EndIf
        \EndIf
    \EndWhile
\EndFor 
\end{algorithmic}
\end{algorithm}
\subsubsection{Option Tree Construction}
The robot's behavior is structured through both the option tree and the subgoal relation tree. The subgoal relation tree is initialized with a starting subgoal at its root, and through this structure, the robot explores multiple possible paths to achieve the task. As the robot progresses through the subgoal tree, options are generated to handle transitions between subgoals.

Each subgoal in the relation tree corresponds to an option in the option tree. In particular, when the robot is at a given subgoal node $g_i$, options $\{o_1, ..., o_{k}\}$ are created to transition between this subgoal and the next subgoal, represented as $g_j$. The structure of the option tree is recursively built as follows:
\begin{equation}
    o^* = \text{create\_child\_option}(o)
    \label{eq:OptionTree}
\end{equation}
where $o^*$ represents a new option generated from the parent option $o$, which is used to connect two subgoals. The termination condition of each option is tied to the completion of its respective subgoal $g_j$, ensuring that the robot moves towards the next goal. As the robot progresses, the option tree grows, connecting various subgoals and forming a more intricate plan for task completion.

This hybrid tree structure allows the robot to connect subgoals using options, enabling it to break down complex tasks into manageable steps. The robot can effectively navigate through different levels of abstraction by leveraging the subgoal relation tree and option tree together.

\subsection{Action Policy}
The action policy is learned under the guidance of a high-level policy over options $\pi_\mathcal{O}$. At the beginning of training, $\pi_\mathcal{O}$ contains only a single global option $o_{z_i}$ for each subgoal $i$. The purpose of the global option is to explore the state space and reach the designated subgoal. This option has an initiation condition that holds across the entire state space $I_{o_{z_i}}(s) = 1$ for all $s$ and terminates only upon reaching the subgoal $\beta_{o_{z_i}} = 1_{g_i}$. During execution, the global option selects primitive actions through its internal policy $\pi_{o_{z_i}}$ until termination.

After the global option $o_{z_i}$ reaches the subgoal $g_i$ a predefined number of times, a new option $o_{g_i}$ is learned from the trajectories collected during these successful attempts. The goal option $o_{g_i}$ shares the same termination condition as the global option, terminating upon reaching the subgoal state. Following this, a new option is established to reach the initiation set of the goal option $o_{g_i}$. By repeating this process, the set of available options in $\pi_\mathcal{O}$ is incrementally expanded as additional skills are discovered.

The learned options are executed over a predefined time horizon $T$. When the robot selects an option $o$ in state $s_t$, the option executes its closed-loop control policy for $\tau$ time steps until either its termination condition is satisfied or the time limit $T$ is reached. Once the option terminates or times out, control is returned to $\pi_\mathcal{O}$, which selects the next option to execute from the resulting state $s_{t+\tau}$.

\begin{figure*}[ht]
    \centering
    \captionsetup[subfigure]{labelformat=empty}
    % First Row
    \subfloat[]{
        \includegraphics[trim=65 40 50 50, clip, width=0.13\textwidth]{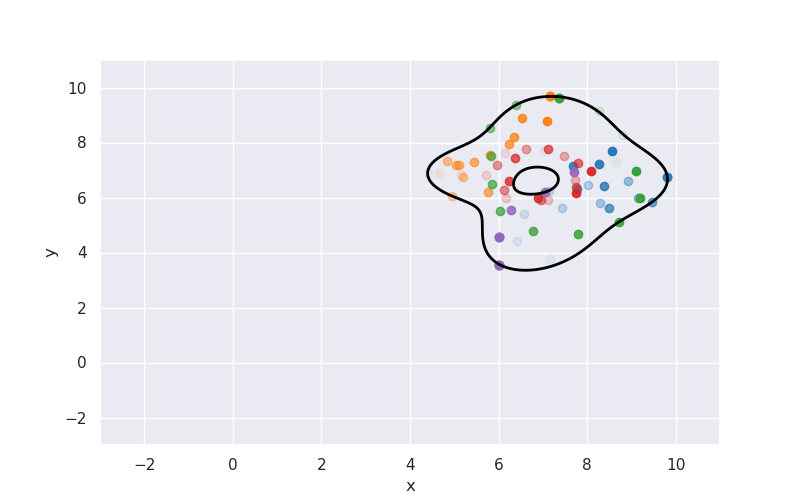}  % Replace with your image path
        \label{fig:caption1}
        
        \includegraphics[trim=65 40 50 50, clip, width=0.13\textwidth]{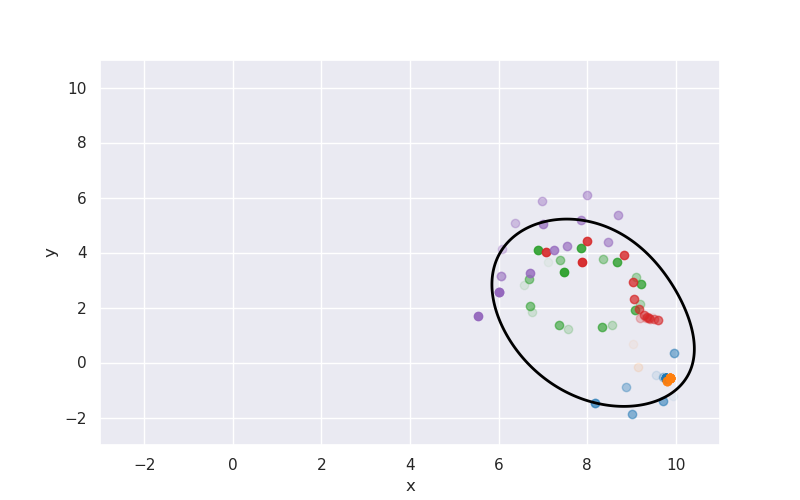}  % Replace with your image path
        \label{fig:caption2}

        \includegraphics[trim=65 40 50 50, clip, width=0.13\textwidth]{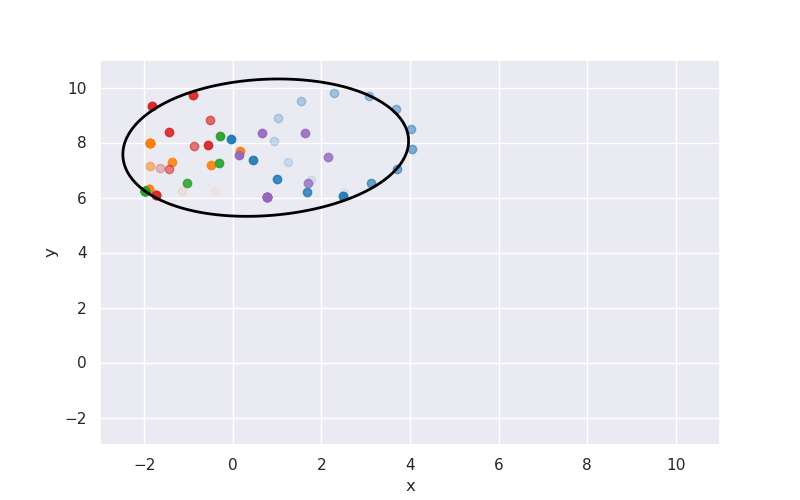}  % Replace with your image path
        \label{fig:caption3}

        \includegraphics[trim=65 40 50 50, clip, width=0.13\textwidth]{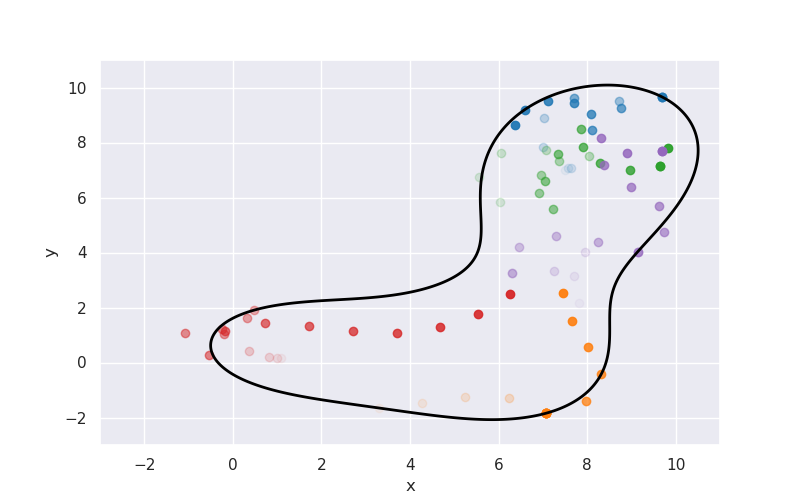}  % Replace with your image path
        \label{fig:caption4}

        \includegraphics[trim=65 40 50 50, clip, width=0.13\textwidth]{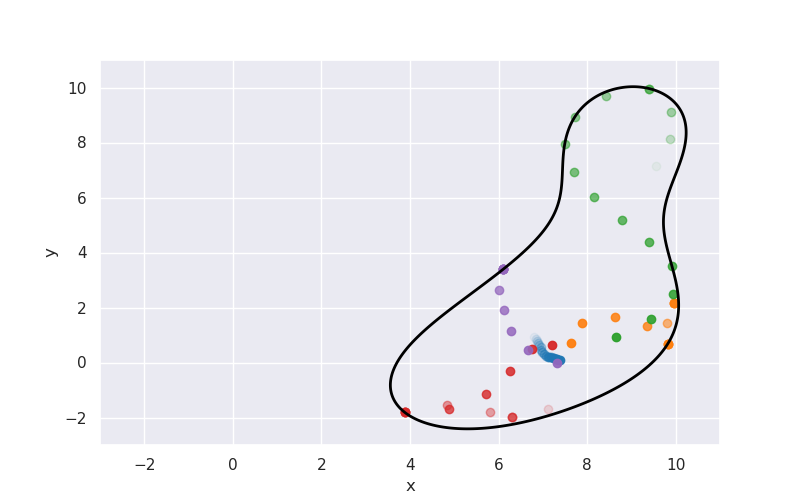}  % Replace with your image path
        \label{fig:caption5}

        \includegraphics[trim=65 40 50 50, clip, width=0.13\textwidth]{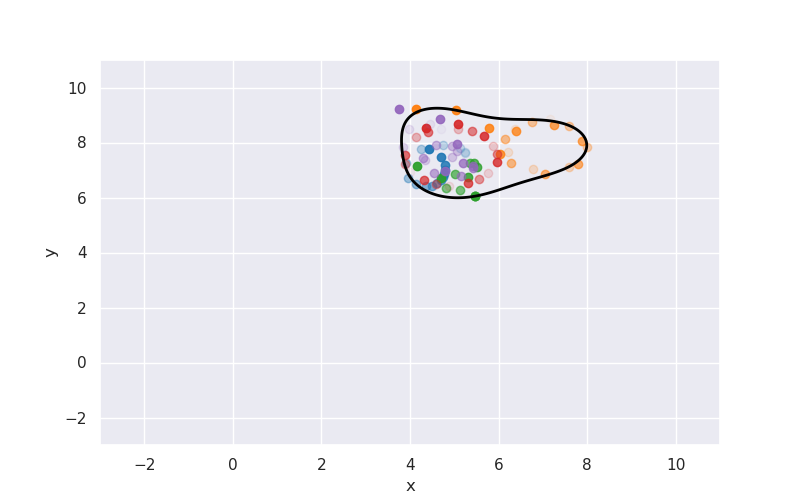}  % Replace with your image path
        \label{fig:caption4}

        \includegraphics[trim=65 40 50 50, clip, width=0.13\textwidth]{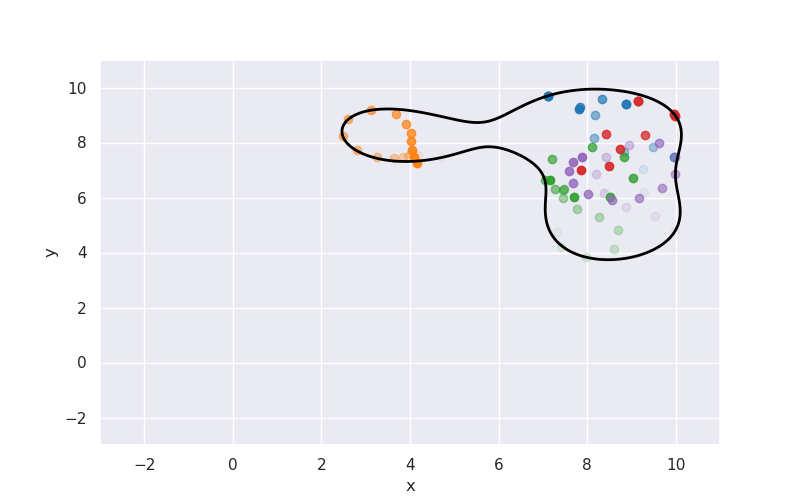}  % Replace with your image path
        \label{fig:caption5}
    }    
    
    % Second Row
    \subfloat[]{
        \includegraphics[trim=50 40 50 50, clip, width=0.13\textwidth]{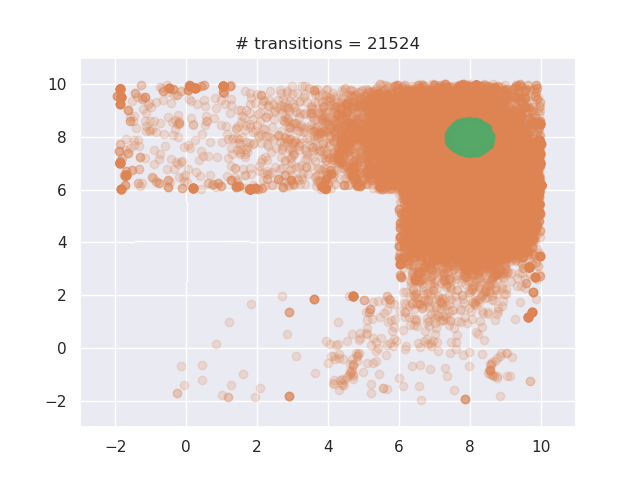}  % Replace with your image path
        \label{fig:caption6}

        \includegraphics[trim=50 40 50 50, clip, width=0.13\textwidth]{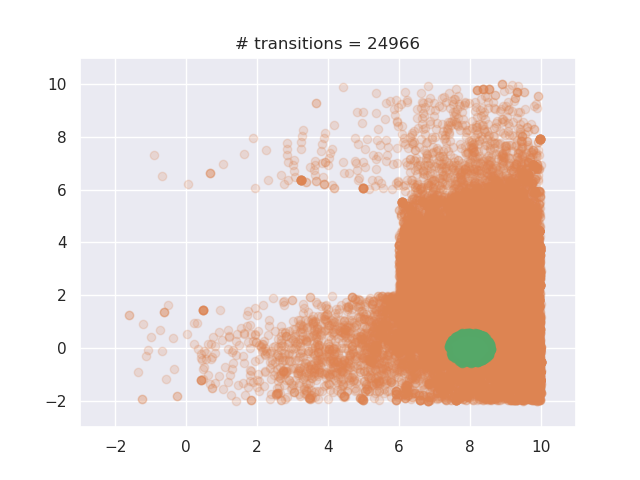}  % Replace with your image path
        \label{fig:caption7}
        \includegraphics[trim=50 40 50 50, clip, width=0.13\textwidth]{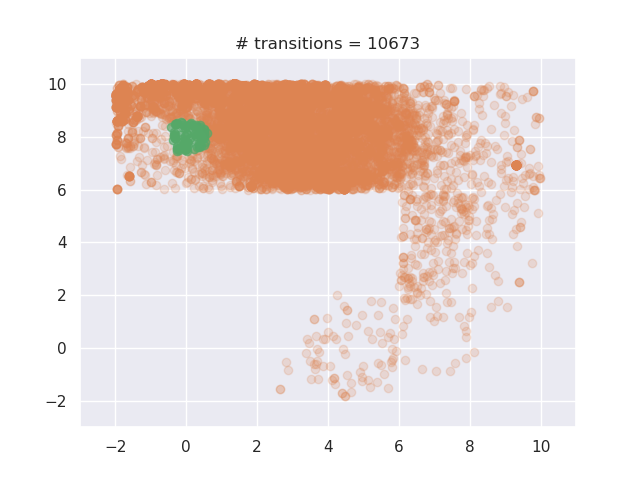}  % Replace with your image path
        \label{fig:caption8}

        \includegraphics[trim=50 40 50 50, clip, width=0.13\textwidth]{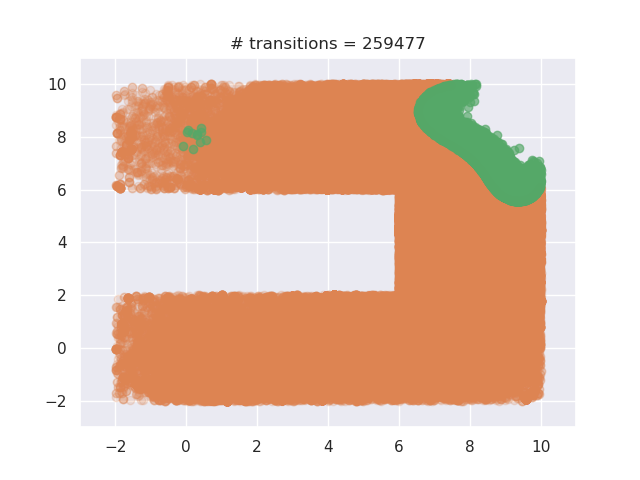}  % Replace with your image path
        \label{fig:caption9}

        \includegraphics[trim=50 40 50 50, clip, width=0.13\textwidth]{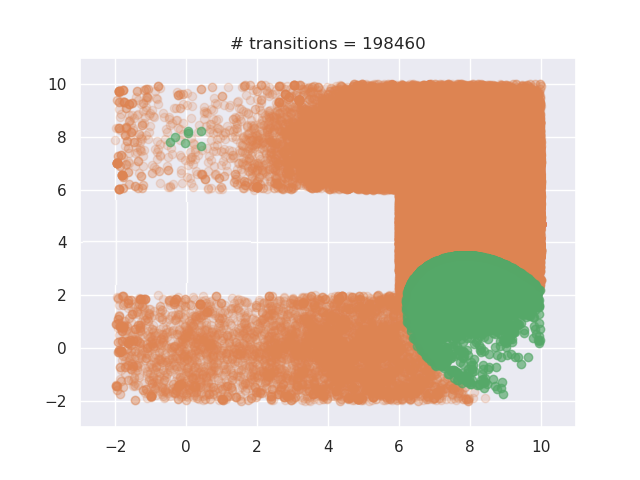}  % Replace with your image path
        \label{fig:caption10}

        \includegraphics[trim=50 40 50 50, clip, width=0.13\textwidth]{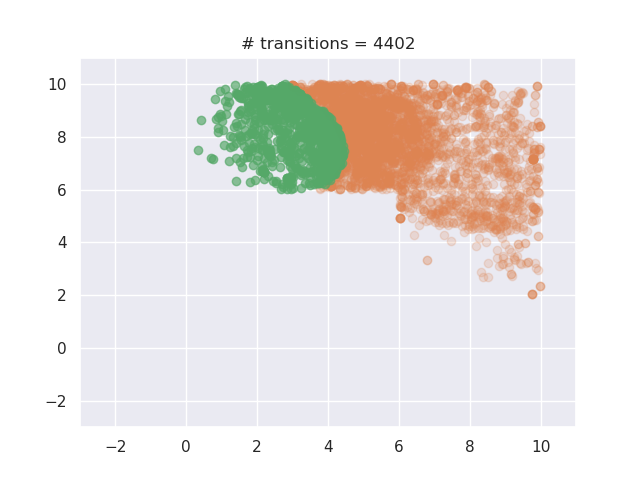}  % Replace with your image path
        \label{fig:caption9}

        \includegraphics[trim=50 40 50 50, clip, width=0.13\textwidth]{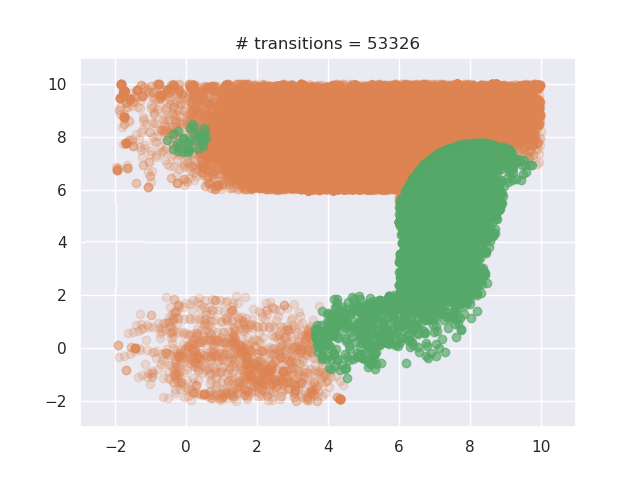}  % Replace with your image path
        \label{fig:caption10}
    }
    
    \includegraphics[trim=60 165 60 165, clip, scale=0.4]{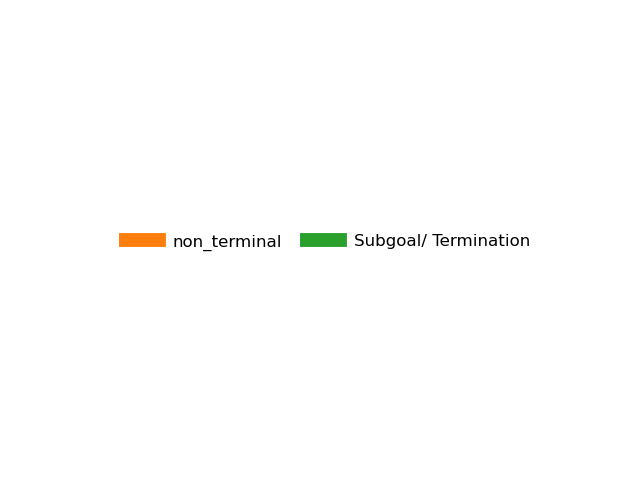}

    \caption{\textbf{Qualitative performance} of the robot in the Point Maze environment. The upper row shows the initial set for each option, illustrating the state space coverage where the option can be executed. The lower row displays the corresponding policy plots, where orange regions indicate areas where the policy continues execution, while green regions signify termination states. The robot follows a structured sequence: first reaching \texttt{subgoal 1} (top-right), then \texttt{subgoal 2} (bottom-right), and finally \texttt{the goal} (top-left).}
    \label{fig:QualitativeExample}
\end{figure*}

\begin{figure}
    \centering
    \includegraphics[width=0.4\linewidth]{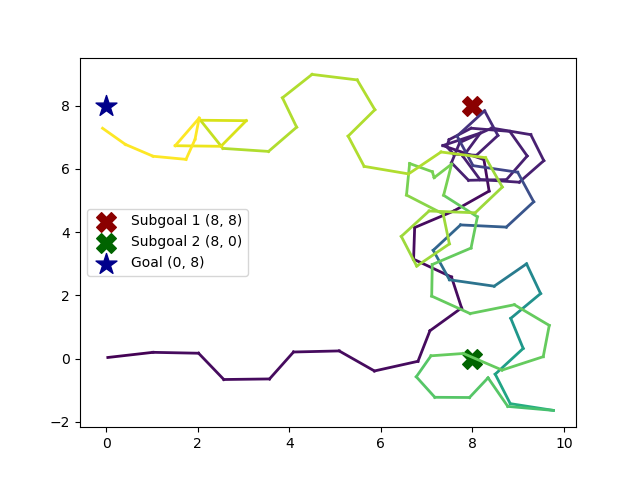}
    \caption{Trajectory of the quantitative example, where the path is colored indicating the change of options. Each segment of the trajectory represents the agent's movement in 2D space while executing a specific option.}
    \label{fig:trajectory}
\end{figure}

\begin{table}[h]
\small
\centering
\caption{Hyperparameters used in the experiments.}
\begin{tabular}{ll}
\toprule
\textbf{Parameter} & \textbf{Value} \\
\midrule
Batch size ($N$) & 64 \\
Discount factor ($\gamma$) & 0.99 \\
Soft update coefficient ($\tau$) & 0.01 \\
Hidden sizes (DDPG) & [400, 300] \\
Hidden layers (DQN) & [32, 32] \\
Critic learning rate & $1 \times 10^{-3}$ \\
Actor learning rate & $1 \times 10^{-4}$ \\
\bottomrule
\end{tabular}
\label{tab:hyperparameters}
\end{table}

\section{Experiments}
We evaluate our algorithm in Mujoco \cite{6386109} framework for RL research. Mujoco is a framework for physical dynamic simulation, often employed to test HRL algorithms. We evaluate our algorithm on four distinct map settings. We apply ChatGPT-o1 for the reasoning. LLM reasoning  Details of the network hyperparameters are provided in Table~\ref{tab:hyperparameters}. 

\subsection{Mujoco}

 \textbf{Four Rooms with Lock and Key}:
    This map is adapted from the Deep Skill Chaining paper \cite{bagaria2019option}, which extends the original Four Rooms environment by introducing a key and lock mechanism. A point robot \cite{duan2016benchmarking} is placed in the Four Rooms setting, where its objective is to first pick up the key and then navigate to the lock, represented by a red sphere in the top-left room. The robot's state space includes its position, orientation, linear velocity, rotational velocity, and a binary \texttt{"has-key"} indicator. The robot must follow the LLM output sequence: \texttt{$<$"retrieve key", "reach the lock"$>$}. If the robot reaches the lock while holding the key, the episode terminates with a sparse reward of 1. Otherwise, the robot incurs a step penalty.
    
\textbf{Point Maze}:
    We extend the Point Maze environment by introducing two subgoal checkpoints. The robot must navigate through these checkpoints sequentially to receive rewards. The LLM output sequence follows the order: \texttt{$<$"retrieve key (green ball)", "Switch on the Remote Control (blue ball)", "go to the door (red ball)"$>$}. The state space and reward structure are designed to emphasize long-term planning and subgoal discovery. The episode terminates once all goals have been traversed. The prompt structure of this task is illustrated in Figure~\ref{fig:prompt}.

 \textbf{Point E-Maze}: We extend the benchmark U-shaped Point-Maze task \cite{bagaria2019option} by introducing two possible starting positions for the robot: one at the top and one at the bottom rungs of the E-shaped maze. Furthermore, we modify the task by incorporating two subgoals in the form of keys, located at the top-right and bottom-right sections of the maze. To reach the final goal, the robot must first collect both keys, requiring it to exhibit strategic planning and skill chaining. The robot can follow either of the two possible LLM outputs:  
\texttt{$<$"retrieve key1 (blue ball)", "retrieve key2 (green ball)", "reach the goal (red ball)"$>$}  
or  
\texttt{$<$"retrieve key2 (green ball)", "retrieve key1 (blue ball)", "reach the goal (red ball)"$>$}.  
This extension challenges the robot to efficiently coordinate subgoal completion before progressing to the final objective.

 \textbf{Tunnel}: The Tunnel environment presents a unique challenge in robot navigation within constrained spaces, evolving the traditional maze by incorporating long, narrow tunnels. These tunnels split into two distinct branches, with a checkpoint located before the branches. At the entrance, the robot must first collect the green key (key1) to unlock the path to the goal. The environment tests the robot's ability to plan and sequence actions strategically. The robot can follow one of the following possible LLM output sequences to complete the task:
\texttt{$<$"go to the checkpoint", "retrieve key1 (green)", "proceed to the goal"$>$},
or
\texttt{$<$"go to the checkpoint", "retrieve key1 (green)", "collect key2 (red)", "proceed to the goal"$>$},
or
\texttt{$<$"go to the checkpoint", "retrieve key2 (red)", "collect key1 (green)", "proceed to the goal"$>$}.
These variations provide the robot with different strategic options, requiring high-level planning.  
%%%
%    \item \textbf{Boss}: This environment is modeled after the BabyAI "Boss" level \cite{babyai_iclr19}. It consists of nine rooms, each containing multiple components. In this task, the robot must first collect two different colored keys—blue and green balls—before reaching the final goal (represented by a red ball). The robot can follow one of the following possible sequences to complete the task:
%\textbf{$<$retrieve the blue key, retrieve the green key, proceed to the goal$>$}, or
%\textbf{$<$retrieve the green key, retrieve the blue key, proceed to the goal$>$}. The environment challenges the robot to effectively plan its route and prioritize key retrieval within the complex room layout.
%%%
\begin{figure*}[ht]
    \centering
    % First Row
    \subfloat[]{
        \makebox[0.23\textwidth][c]{
        \includegraphics[width=0.14\textwidth]{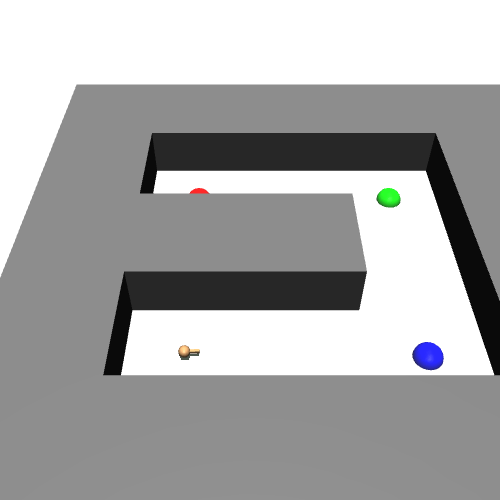}  % Replace with your image path
        }
        \label{fig:caption1}
    }
    \subfloat[]{
        \makebox[0.23\textwidth][c]{ % 0.23 width box, centered content
        \includegraphics[trim=0 0 0 50, clip,width=0.14\textwidth]{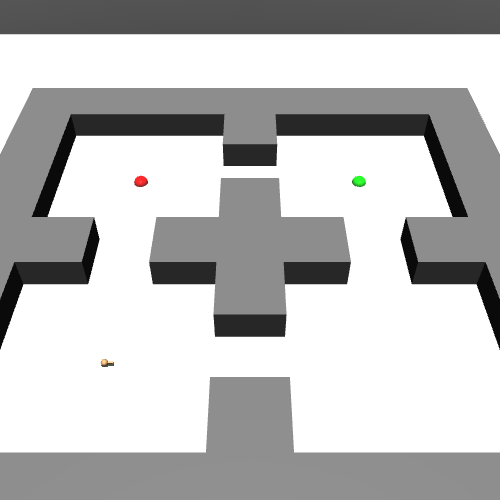}  % Replace with your image path
        }
        \label{fig:caption2}
    }
    \subfloat[]{
        \makebox[0.23\textwidth][c]{
        \includegraphics[width=0.14\textwidth]{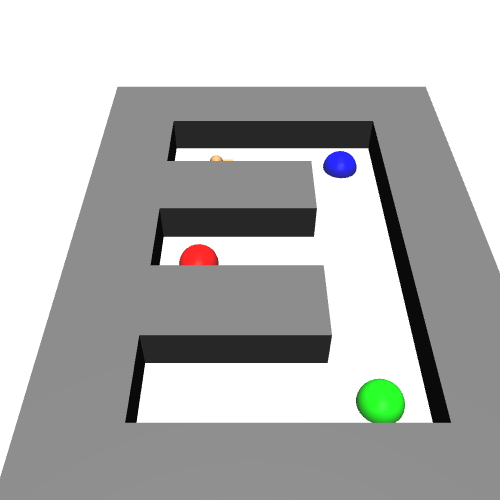}  % Replace with your image path
        }
        \label{fig:caption3}
    }
    \subfloat[]{
        \makebox[0.23\textwidth][c]{
        \includegraphics[width=0.14\textwidth]{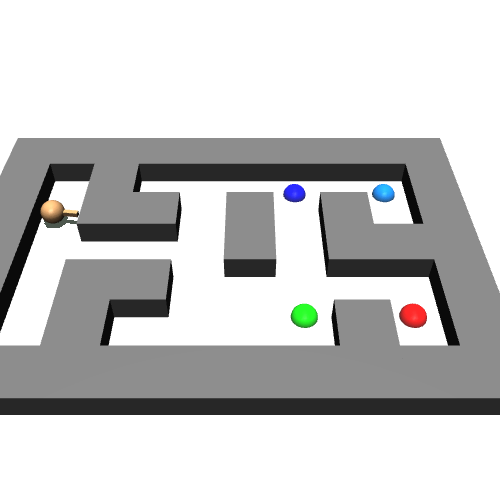}  % Replace with your image path
        }
        \label{fig:caption4}
    }

    % Second Row
    \subfloat[ ]{
        \makebox[0.23\textwidth][c]{
        \includegraphics[trim=0 0 50 40, clip,width=0.23\textwidth]{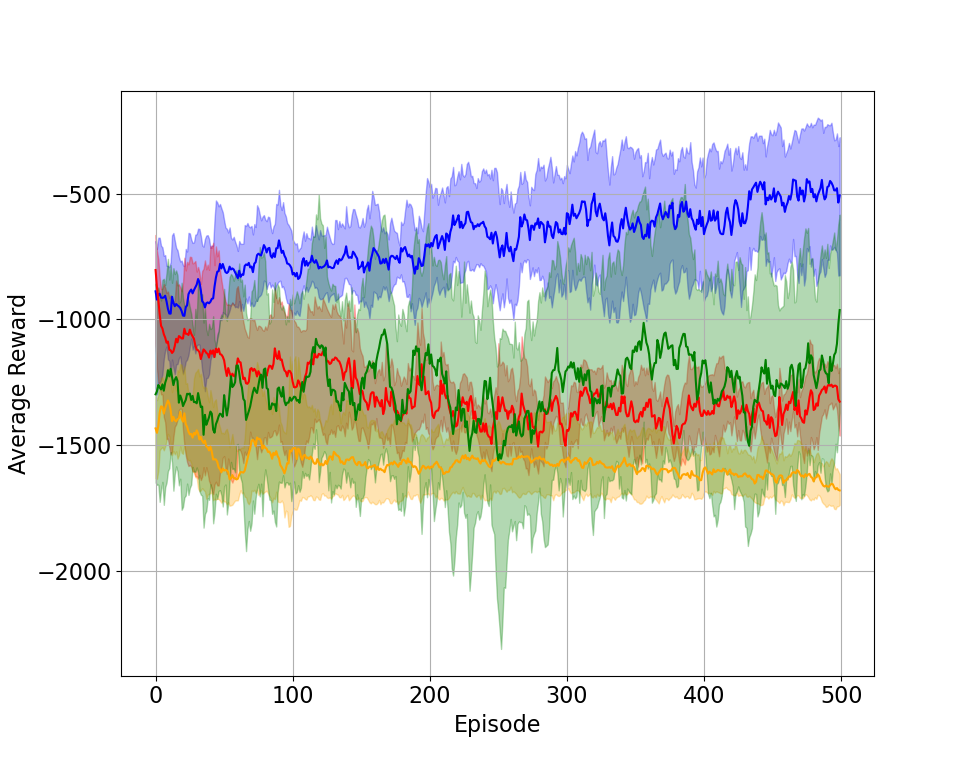}  % Replace with your image path
        }
        \label{fig:caption6}
    }
    \subfloat[]{
        \makebox[0.23\textwidth][c]{
        \includegraphics[trim=0 0 50 0, clip, width=0.22\textwidth]{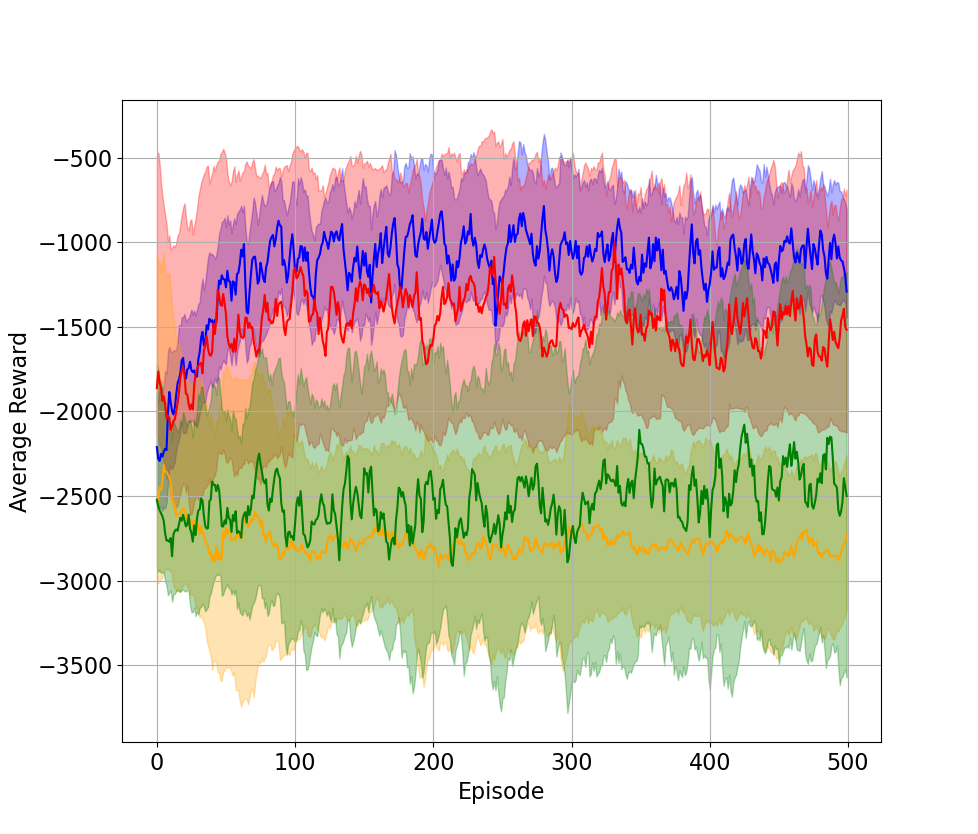}  % Replace with your image path
        }
        \label{fig:caption7}
    }
    \subfloat[]{
        \makebox[0.23\textwidth][c]{
        \includegraphics[trim=0 0 50 50, clip, width=0.23\textwidth]{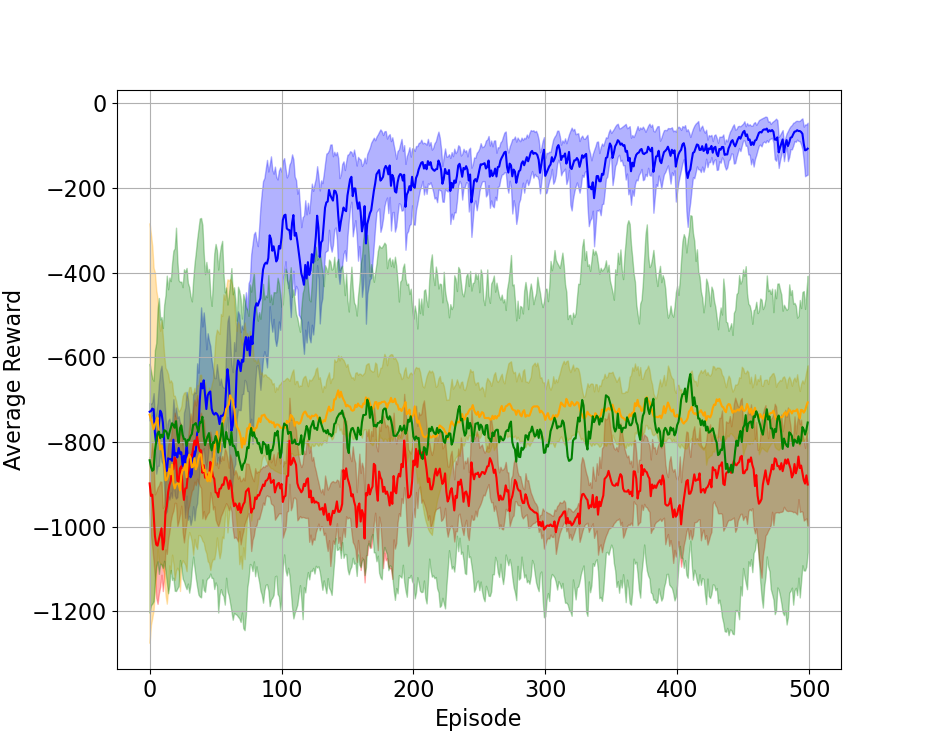}  % Replace with your image path
        }
        \label{fig:caption8}
    }
    \subfloat[]{
        \makebox[0.23\textwidth][c]{
        \includegraphics[width=0.23\textwidth]{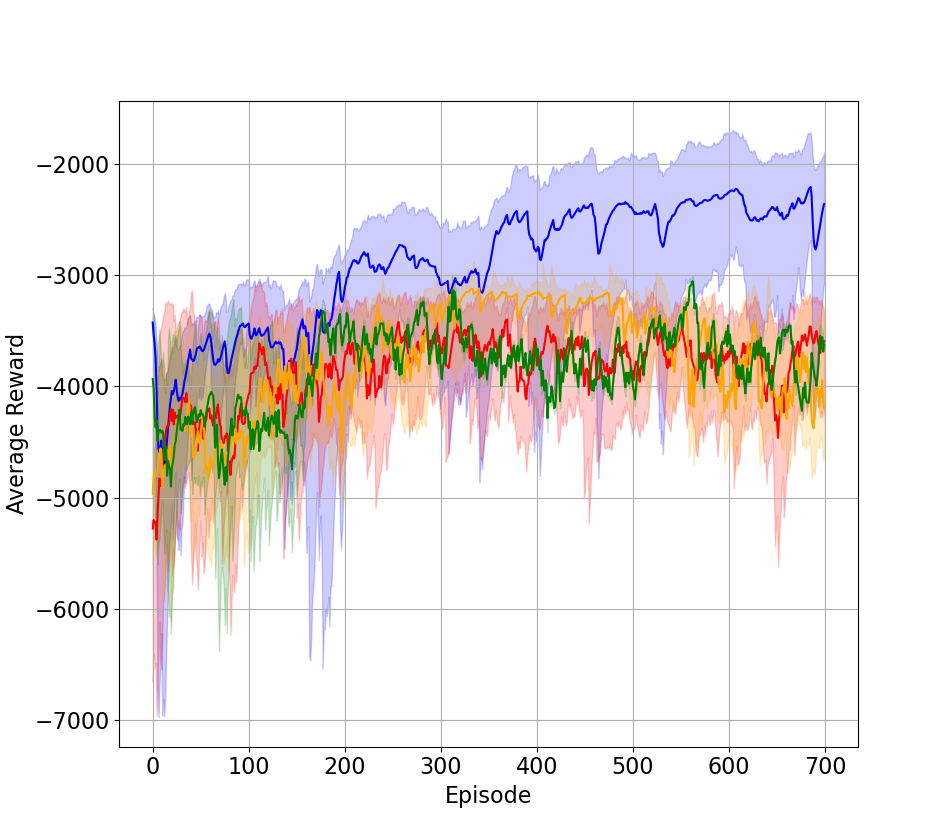}  % Replace with your image path
        }
        \label{fig:caption9}
    }

    \includegraphics[trim=0 7 0 0, clip, scale=0.4]{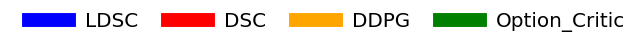}
    \caption{(Upper) Environment setup and (Bottom) average reward for each task: (a) and (e) are Maze, (b) and (f) are Four Rooms, (c) and (g) are E-Maze, and (d) and (h) are Tunnel.}
    \label{fig:two_row_five_col}
\end{figure*}

\begin{table*}[ht]
\centering
\resizebox{\textwidth}{!}{ 
\begin{tabular}{|c|c|c|c|c|c|c|c|c|}
\hline 
\multirow{2}{*}{Map} & \multicolumn{2}{c|}{\textbf{DSC}} & \multicolumn{2}{c|}{\textbf{DDPG}} & \multicolumn{2}{c|}{\textbf{Option-Critic}} & \multicolumn{2}{c|}{\textbf{LDSC (Ours)}} \\ \cline{2-9}
                     & Success Rate $\uparrow$ & Time (s) $\downarrow$ & Success Rate $\uparrow$ & Time (s) $\downarrow$ & Success Rate $\uparrow$ & Time (s) $\downarrow$ & Success Rate $\uparrow$ & Time (s) $\downarrow$ \\ \hline
\textbf{Maze}       & 0\% $\pm$ 0\%  & 1063 $\pm$ 274  & 0\% $\pm$ 0\% & 1187 $\pm$ 116  & 0\% $\pm$ 0\%  & 1030 $\pm$ 147  & \textbf{100\%} $\pm$ \textbf{0\%}  & \textbf{485} $\pm$ \textbf{174}  \\ \hline
\textbf{FourRoom}   & 86\% $\pm$ 8\% & 1035 $\pm$ 479  & 0\% $\pm$ 0\% & 1331 $\pm$ 355  & 0\% $\pm$ 0\%  & 1181 $\pm$ 500  & \textbf{95\%} $\pm$ \textbf{2\%}   & \textbf{678} $\pm$ \textbf{208}  \\ \hline
\textbf{E-Maze}     & 0\% $\pm$ 0\%  & 1345 $\pm$ 145  & 0\% $\pm$ 0\% & 1344 $\pm$ 83   & 0\% $\pm$ 0\%  & 960 $\pm$ 306   & \textbf{100\%} $\pm$ \textbf{0\%}  & \textbf{86.5} $\pm$ \textbf{12.5} \\ \hline
\textbf{Tunnel}     & 0\% $\pm$ 0\%  & 1368 $\pm$ 370  & 0\% $\pm$ 0\% & 1527 $\pm$ 527  & 0\% $\pm$ 0\%  & 1349 $\pm$ 349  & \textbf{81.8\%} $\pm$ \textbf{3.6\%} & \textbf{906} $\pm$ \textbf{306}   \\ \hline
\end{tabular}
}
\caption{Comparison of different methods in terms of success rate and completion time across various maps.}
\label{tab:methods_comparison}
\end{table*}

\subsection{Baselines}
We have considered the following baseline algorithms. \textbf{Deep Skill Chaining (DSC)} \cite{bagaria2019option}: A HRL method that decomposes complex tasks into a sequence of sub-skills, enabling more efficient learning and planning. \textbf{Deep Deterministic Policy Gradient (DDPG)} \cite{lillicrap2015continuous}: A model-free, off-policy actor-critic algorithm designed for continuous action spaces, leveraging experience replay and target networks for stable training.
\textbf{Option-Critic} \cite{bacon2017option}: A framework that combines the option-based hierarchy with a policy gradient method, allowing the robot to learn both policies over options and policies within options.

\subsection{Evaluation}

\subsubsection{Qualitative Example}

To assess the qualitative performance of the proposed LDSC framework, we present visual examples that illustrate its behavior in a Maze environment. As shown in Figure \ref{fig:QualitativeExample}, LDSC effectively learns high-level abstract options that enable structured and hierarchical navigation toward subgoals and the final goal. The first three options demonstrate localized initiation sets and policies that guide the robot to key subgoal locations, including the top-right, top-left, and bottom-right corners of the maze. Beyond these localized strategies, LDSC also generates bridging options that facilitate transitions between subgoals and the final goal. Notably, as shown in Figure \ref{fig:trajectory}, the trajectory of the quantitative example demonstrates how the agent transitions between different options, with the path colored according to the active option, demonstrating its capability to form temporally extended and goal-directed policies.

This hierarchical decomposition of the navigation task into smaller, modular subproblems significantly improves the efficiency of planning and execution. By leveraging learned options, the agent effectively reduces unnecessary exploration, ensuring more directed and optimized movement between subgoals and the final destination. The ability to abstract and generalize decision-making at different levels of granularity further highlights the strength of LDSC in solving complex, long-horizon navigation tasks.

\subsubsection{Quantitative Comparison}

We evaluate the robot's performance by assessing its task completion rates and the time taken to complete each task based on human-provided goals. The corresponding visual representations are shown in Figures \ref{fig:two_row_five_col} (Upper), while the performance results are depicted in Figures \ref{fig:two_row_five_col} (Bottom).

Our approach, LDSC, demonstrates superior performance compared to all baseline algorithms without requiring additional training time, as the LLM processing is completed beforehand. Notably, while DSC performs better than the baseline methods, LDSC achieves a significant margin of improvement across all tasks. On average, LDSC outperforms baseline methods by \textbf{55.9\%}, reduces task completion time by \textbf{53.1\%}, and improves success rates by \textbf{72.7\%}, ensuring more efficient and reliable execution. 

The key advantage of LDSC lies in its LLM-driven hierarchical decomposition, where the problem is broken down into smaller, localized subproblems using reasoning capabilities from a language model. By leveraging LLM-based reasoning, LDSC can infer logical subgoal structures, effectively segmenting the task into meaningful intermediate steps. This hierarchical structure significantly reduces the search space, thereby enhancing exploration efficiency. Since the agent can now focus on reaching subgoals rather than exploring the entire state space at once, it requires fewer interactions to discover optimal behaviors. As a result, the learning speed is significantly accelerated, and the agent exhibits a more robust and stable performance across different maze configurations. This is particularly evident in the Point E-Maze setting, where the introduction of multiple subgoals increases task complexity, yet LDSC still outperforms other methods by a substantial margin. We also evaluate the robot’s performance in a complex environment, Tunnel, which requires long-term planning. LDSC significantly outperforms baseline methods, demonstrating superior efficiency and adaptability in challenging scenarios.

These results highlight the effectiveness of LDSC in leveraging high-level abstraction and LLM-driven reasoning to improve long-horizon planning, ultimately leading to higher task success rates and more efficient navigation in complex environments.

\section{Conclusion}
In this work, we introduced LDSC, a semantic HRL method that leverages LLMs for subgoal reasoning and decomposition. Our approach effectively addresses key challenges in exploration efficiency, policy generalization, and option reusability in complex, multi-task reinforcement learning settings. By integrating LLM-driven reasoning, LDSC constructs structured subgoal hierarchies, significantly improving task decomposition and learning efficiency. This is achieved without increasing overall training time, as the LLM operates only during the pre-processing stage. Through extensive experiments in diverse maze environments, we demonstrated that LDSC outperforms the existing RL approaches DSC, DDPG, and Option-Critic methods. Our results show that LDSC achieves a substantial improvement in task success rates, task completion time, and overall learning efficiency, highlighting its robustness and adaptability across different environments.

The significance of LDSC extends beyond the current experiments. Future research can explore scaling LDSC to high-dimensional robotic tasks and further refining semantic reasoning mechanisms to enhance adaptability. Additionally, investigating multi-agent collaboration and lifelong learning extensions for LDSC could open new directions in RL.

Overall, LDSC provides a compelling framework for hierarchical decision-making, offering a scalable and interpretable approach to RL in dynamic and uncertain environments.
% In the unusual situation where you want a paper to appear in the
% references without citing it in the main text, use \nocite
\nocite{langley00}

\bibliography{paper}
\bibliographystyle{IEEEtran}

%%%%%%%%%%%%%%%%%%%%%%%%%%%%%%%%%%%%%%%%%%%%%%%%%%%%%%%%%%%%%%%%%%%%%%%%%%%%%%%
%%%%%%%%%%%%%%%%%%%%%%%%%%%%%%%%%%%%%%%%%%%%%%%%%%%%%%%%%%%%%%%%%%%%%%%%%%%%%%%
% APPENDIX

%%%%%%%%%%%%%%%%%%%%%%%%%%%%%%%%%%%%%%%%%%%%%%%%%%%%%%%%%%%%%%%%%%%%%%%%%%%%%%%
%%%%%%%%%%%%%%%%%%%%%%%%%%%%%%%%%%%%%%%%%%%%%%%%%%%%%%%%%%%%%%%%%%%%%%%%%%%%%%%

%\Function{execute\_option}{$o_t$}
%    \State $t_0 = t$
%    \State $T$ is the option’s episodic time budget
%    \State $\pi_{o_t}$ is the option’s internal policy
%    \While{not $\beta_{o_t}(s_t)$ \textbf{and} $t < T$}
%        \State $a_t = \pi_{o_t}(s_t; \theta_{o_t})$
%        \State $r_t, s_{t+1} = env.step(a_t)$
%        \State $s_t = s_{t+1}$
%        \State $t = t + 1$
%    \EndWhile
%    \State $\tau = t$ // duration of option execution
%    \State \Return $r_{t_0 : t_0 + \tau}, s_{t_0 + \tau}$
%\EndFunction

%%%%%%%%%%%%%%%%%%%%%%%%%%%%%%%%%%%%%%%%%%%%%%%%%%%%%%%%%%%%%%%%%%%%%%%%%%%%%%%

%Hierarchical Continual Reinforcement Learning via Large Language Model 

%Deep Skill Chaining
%Effectively Learning Initiation Sets in HRL
%Hierarchical contextual RL via LLM \\

%Option Critic
%Transformer Critic

%%%%%%%%%%%%%%%%%%%%%%%%%%%%%%%%%%%%%%%%%%%%%%%%%%%%%%%%%%%%%%%%%%%%%%%%%%%%%%%

%\subsubsection{Multiple Tasks}
%We evaluate the robot's ability to handle multiple tasks at once, measuring its proficiency in managing diverse goals and the impact of concurrent task learning on overall efficiency.

\end{document}